\documentclass{article}

\usepackage[preprint]{neurips_2023}

\usepackage[utf8]{inputenc} 
\usepackage[T1]{fontenc}    

\usepackage{tgpagella}

\usepackage{hyperref}       
\usepackage{url}            
\usepackage{booktabs}       
\usepackage{amsfonts}       
\usepackage{nicefrac}       
\usepackage{microtype}      
\usepackage[table,xcdraw]{xcolor}         

\usepackage{hyperref}
\usepackage{url}
\usepackage{graphicx}
\usepackage{scalerel}

\usepackage{booktabs}
\usepackage{multirow}
\usepackage{makecell}
\usepackage{adjustbox}

\usepackage{soul}

\definecolor{darkblue}{rgb}{0, 0, 0.5}
\hypersetup{
    colorlinks=true,
    linkcolor=darkblue,
    citecolor=darkblue,
    pdfpagemode=FullScreen,
    }

\usepackage{booktabs}
\usepackage[ruled,vlined]{algorithm2e}
\definecolor{commentcolor}{RGB}{110,154,155}   

\usepackage{listings} 

\definecolor{codegreen}{rgb}{0,0.6,0}
\definecolor{codegray}{rgb}{0.5,0.5,0.5}
\definecolor{codepurple}{rgb}{0.58,0,0.82}
\definecolor{backcolour}{rgb}{0.95,0.95,0.92}

\lstdefinestyle{mystyle}{
    backgroundcolor=\color{backcolour},   
    commentstyle=\color{codegreen},
    keywordstyle=\color{magenta},
    numberstyle=\tiny\color{codegray},
    stringstyle=\color{codepurple},
    basicstyle=\ttfamily\footnotesize,
    breakatwhitespace=false,         
    breaklines=true,                 
    captionpos=b,                    
    keepspaces=true,                 
    numbers=left,                    
    numbersep=5pt,                  
    showspaces=false,                
    showstringspaces=false,
    showtabs=false,                  
    tabsize=2
}
\newcommand{\logo}{\makebox[0pt][l]{\hspace{0pt}\raisebox{-0.8ex}{\includegraphics[height=25pt]{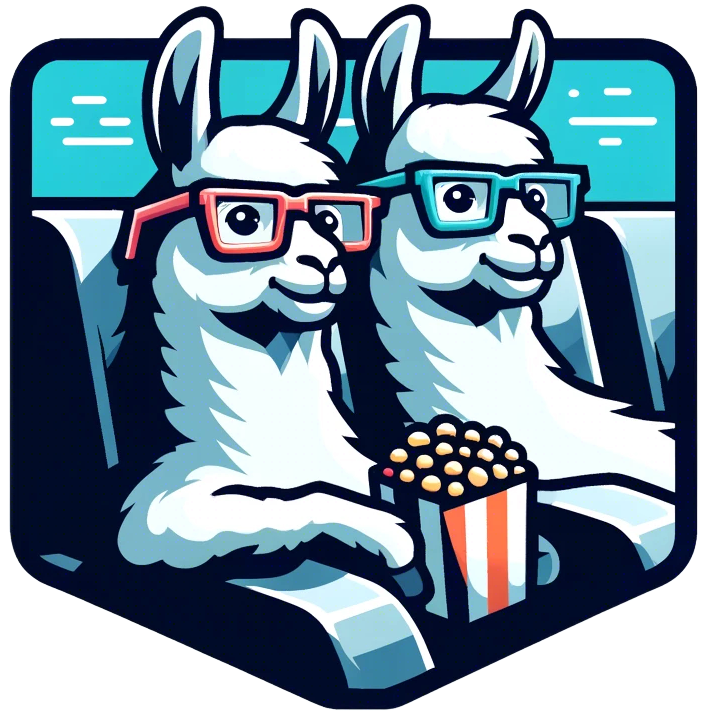}}}}

\title{\logo \ \ \ \ \ \ \ VideoLLaMA 2\\ Advancing Spatial-Temporal Modeling and Audio Understanding in Video-LLMs}

\author{%
  Zesen Cheng\thanks{\ \ ZC, SL, HZ, YX, and XL contributed equally to this project.} , Sicong Leng$^*$, Hang Zhang$^*$, Yifei Xin$^*$, Xin Li$^*$,\\
  \textbf{Guanzheng Chen, Yongxin Zhu, Wenqi Zhang, Ziyang Luo, Deli Zhao, Lidong Bing} \\ \\
  DAMO Academy, Alibaba Group \\ \\
  \url{https://github.com/DAMO-NLP-SG/VideoLLaMA2} \\
}

\usepackage{multirow}
\usepackage{multicol}
\usepackage{bbm}
\usepackage{arydshln}

\usepackage{subcaption}

\DeclareMathAlphabet\mathbfcal{OMS}{cmsy}{b}{n}

\definecolor{cgreen}{HTML}{39b54a}  
\definecolor{cyellow}{HTML}{FFC000}
\definecolor{cred}{HTML}{A10035}
\definecolor{cpurple}{HTML}{7030A0}
\definecolor{alicegreen}{HTML}{E7F5E1}
\definecolor{alicedarkgreen}{HTML}{A6DA8E}
\definecolor{lightpurple}{HTML}{8C88C1}
\definecolor{plum}{HTML}{9C276A}
\definecolor{darkpurple}{HTML}{3C1B61}

\usepackage{pifont}
\newcommand{\cmark}{\textcolor{purple}{\ding{52}}}%

\newcommand{\listnumber}[1]{\textbf{\color{violet}{#1}}}

\usepackage{tabulary}
\newcolumntype{I}{!{\vrule width 1pt}}

\newcolumntype{x}[1]{>{\centering\arraybackslash}p{#1pt}}
\newcolumntype{y}[1]{>{\raggedright\arraybackslash}p{#1pt}}
\newcolumntype{z}[1]{>{\raggedleft\arraybackslash}p{#1pt}}
\newlength\savewidth

\makeatletter
\newcommand{\thickhline}{%
	\noalign {\ifnum 0=`}\fi \hrule height 1pt
	\futurelet \reserved@a \@xhline
}
\makeatother

\newcommand{\one}{\textcolor{violet}{\ding{182}}}
\newcommand{\two}{\textcolor{violet}{\ding{183}}}
\newcommand{\three}{\textcolor{violet}{\ding{184}}}
\newcommand{\four}{\textcolor{violet}{\ding{185}}}

\usepackage{fdsymbol}
\newcommand{\official}{\textcolor{plum}{$^{\varheartsuit}$}}
\newcommand{\leaderboard}{$^{\diamondsuit}$}
\newcommand{\reproduce}{\textcolor{darkpurple}{$^{\spadesuit}$}}

\usepackage{CJKutf8}

\begin{document}

\maketitle

\begin{abstract}
In this paper, we present VideoLLaMA 2, a set of Video Large Language Models (Video-LLMs) designed to enhance spatial-temporal modeling and audio understanding in video and audio-oriented tasks.
Building upon its predecessor, VideoLLaMA 2 incorporates a tailor-made \textit{Spatial-Temporal Convolution (STC) connector}, which effectively captures the intricate spatial and temporal dynamics of video data. Additionally, we integrate an \textit{Audio Branch} into the model through joint training, thereby enriching the multimodal understanding capabilities of the model by seamlessly incorporating audio cues.
Comprehensive evaluations on multiple-choice video question answering~(MC-VQA), open-ended video question answering~(OE-VQA), and video captioning~(VC) tasks demonstrate that VideoLLaMA 2 consistently achieves competitive results among open-source models and even gets close to some proprietary models on several benchmarks.
Furthermore, VideoLLaMA 2 exhibits reasonable improvements in audio-only and audio-video question-answering (AQA \& OE-AVQA) benchmarks over existing models.
All models are public to facilitate further research.

\end{abstract}

\section{Introduction}
In recent years, the field of artificial intelligence (AI) has achieved significant advancements~\citep{openai2023gpt4tr,team2023gemini,anthropic2024claude3}, profoundly transforming industries and societal functions across the board. 
Models capable of image recognition~\citep{bai2023qwen,dong2024internlm,chen2023internvl,liu2024visual,chen2023sharegpt4v,lin2023sphinx,young2024yi} and photorealistic image generation~\citep{esser2024scaling,saharia2022photorealistic,ramesh2021zero} have approached near-human capabilities, catalyzing major breakthroughs in sectors such as medical imaging~\citep{li2024llava,tu2024towards} and autonomous driving~\citep{xu2023drivegpt4,jin2023adapt}. 
Despite these successes, the domain of video understanding and generation~\citep{kondratyuk2023videopoet,menapace2024snap,kondratyuk2023videopoet,videoworldsimulators2024} remains relatively nascent. Unlike static images, videos incorporate temporal dynamics and synchronous audio streams, significantly enriching the information content.
This integration of continuous audio-visual data complicates extracting and interpreting meaningful patterns, amplifying data complexity and introducing unique computational challenges.

While Image Large Language Models (Image-LLMs)~\citep{alayrac2022flamingoav,li2023blip2bl,zhu2023minigpt,liu2024visual,ye2023mplugowl,bai2023qwen,chen2023internvl,dong2024internlm} processing static images have matured with impressive capabilities, Video Large Language Models (Video-LLMs)~\citep{li2023videochat,zhang2023video,maaz2023video,lin2023video,wang2024internvideo2,liu2024world} lag notably behind due to inherent complexities.
The primary challenge in video understanding lies in the temporal dynamics—recognizing visual patterns while understanding changes over time and correlating these with synchronous audio inputs. 
These temporal dynamics complicate the accurate prediction of future states and the understanding of complex scenarios, such as interactions among multiple entities or subtle environmental changes.
Moreover, the integration of audio with visual data, essential for comprehensive understanding, remains underdeveloped in current models, limiting their effectiveness. 

Current Video-LLMs are constrained by several limitations that affect their performance and utility. Firstly, these models often struggle with effectively processing temporal dynamics due to their limited capabilities in fusing features across different frames. This results in a failure to fully capitalize on the available temporal information, hindering their ability to predict future events accurately based on past and present data. Secondly, the integration of audio streams is frequently overlooked, despite audio being a rich source of contextual cues that are vital for a complete scene understanding. This neglect leads to a significant gap in models' ability to perform comprehensive multimodal analyses. 
These limitations illustrate the need for more advanced Video-LLMs that can handle the complexities of multimodal video data without compromising processing efficiency or contextual integrity.

This technical report unveils the VideoLLaMA 2, a set of generalist Video-LLMs designed to enhance video-language understanding by integrating and interpreting the complex interplay of visual and auditory signals. 
Built on the technical foundations established by earlier models~\citep{radford2021learning,radosavovic2020designing,chen2023beats,cha2023honeybee,jiang2023mistral}, VideoLLaMA 2 delivers a system capable of not only comprehending but also articulating the rich narratives inherent in video content.
The model’s robust performance is anchored by its ability to effectively process temporal dynamics, achieved through the implementation of our specially designed \textit{Spatial-Temporal Connector} module. This allows VideoLLaMA 2 to excel in various video-language tasks, from video captioning to complex question answering, demonstrating a profound comprehension of video content.
Further contributing to its effectiveness is the joint-trained \textit{Audio Branch}, which enhances VideoLLaMA 2’s capacity for advanced audio-visual integration. This feature ensures that audio data, often underutilized in video language models, significantly bolsters the interpretative depth, capturing subtle cues lost in visual-only analyses. 

These technological enhancements make VideoLLaMA 2 a pivotal development in video-language analytics, setting a new standard for the capabilities of intelligent video understanding systems. The subsequent sections of this report will delve into the technical architecture of the model, explore the innovative methodologies employed, and present a detailed evaluation of its performance, illustrating its superiority over existing models.

\begin{figure}[t]
\centering
\includegraphics[width=1.0\textwidth]{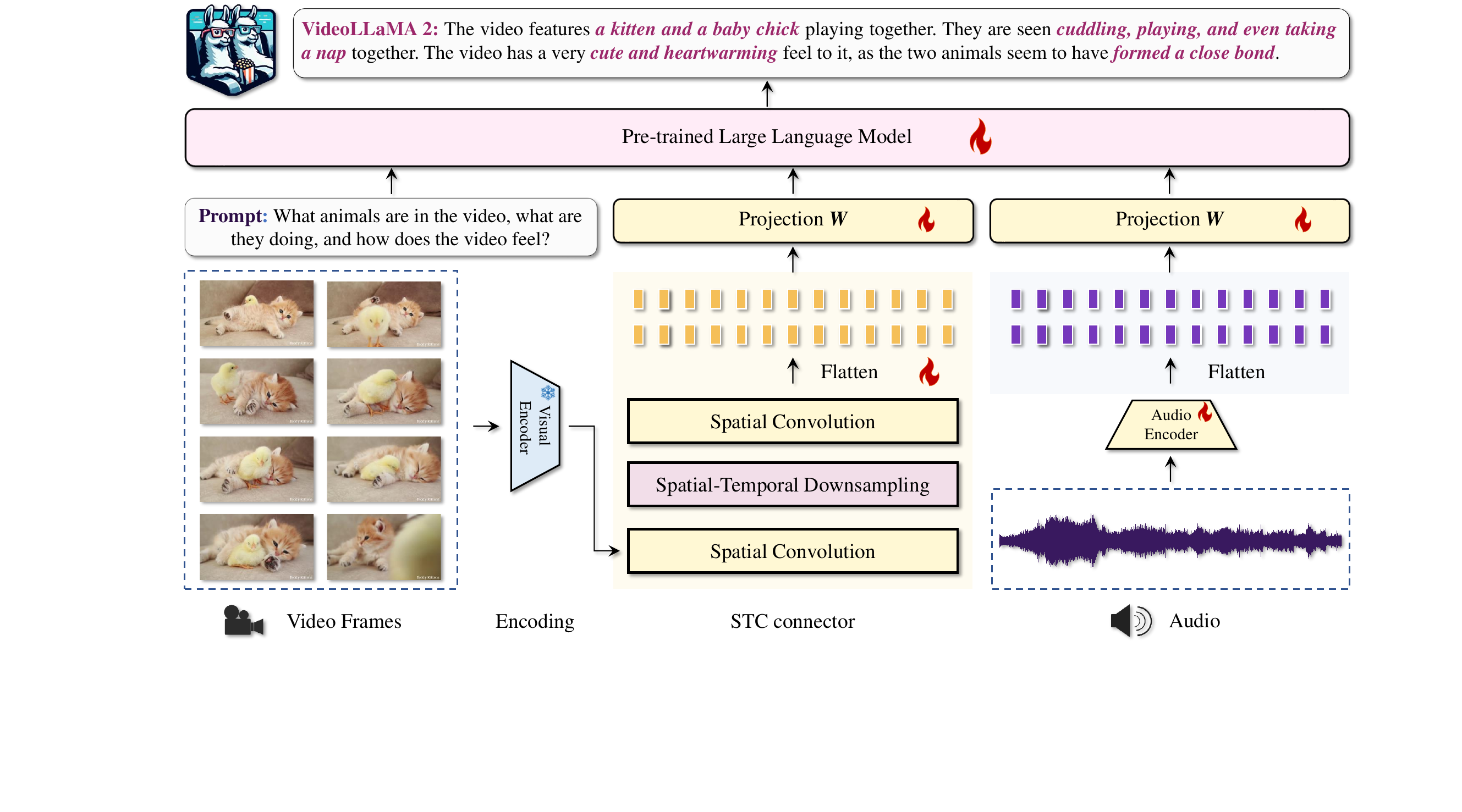} 
\caption{\textbf{The overall pipeline} of VideoLLaMA 2. For the vision-language branch, the video frames are encoded into features frame by frame, processed through our STC connector, and then these features are fed into a large language model to generate responses based on text prompts. For the audio-language branch, audio signals are first transformed into log mel spectrograms, which are then encoded to extract auditory features. These features are then processed through a multilayer perceptron (MLP) block to align the audio modalities with the large language model.
}
\label{fig:overall_pipeline}
\end{figure}

\section{Method}

\subsection{Architecture} 
As depicted in Figure~\ref{fig:overall_pipeline}, VideoLLaMA 2 adheres to the design principle established in its previous version (i.e., VideoLLaMA~\citep{zhang2023video}), which integrates a dual-branch framework comprised of a Vision-Language Branch and an Audio-Language Branch. Both branches operate independently, connecting pre-trained visual and audio encoders to an instruction-finetuned large language model in a modular fashion. This modality-specific independence of the visual and audio branches, with cross-modal interactions occurring solely within the highly capable language model, not only allows streamlined training by preserving the integrity of individual modal inputs, but also facilitates future expansions and adaptations.

\paragraph{Vision-Language Branch} Although video modality is our main focus, we choose image-level CLIP~(ViT-L/14)~\citep{radford2021learning} as our vision backbone. The main reason is that image encoders are compatible with arbitrary frame sampling strategies and enable a more flexible frame-to-video feature aggregation scheme, as observed in~\cite{xue2023clip}. During training and inference, we adopt a consistent frame sampling approach that extracts a fixed number of frames from each video. Each frame undergoes padding and resizing to a standardized 336x336 dimension. The preprocessed frames are then fed into the image encoder. Instead of the Q-former in VideoLLaMA 1~\citep{zhang2023video}, we propose a Spatial-Temporal Convolution Connector (STC Connector) in VideoLLaMA 2 for spatial-temporal representation learning. The STC Connector could preserve spatial and temporal local details more effectively than the Q-former while not producing a large number of video tokens. A detailed exploration of the working mechanism and advantages of the STC Connector is provided in Section~\ref{sec.connector}.

\paragraph{Audio-Language Branch} With an established foundation in the visual domain, our exploration extends into the auditory realm to enhance the multimodal capabilities of VideoLLaMA. Initially, audio signals undergo a preprocessing step where they are transformed into fbank spectrograms with 128 frequency bins. To effectively harness these preprocessed audio signals, we integrate BEATs \citep{chen2023beats}, a cutting-edge audio encoder, known for its exceptional ability to capture detailed audio features and temporal dynamics. These features are then processed through a MLP block with two linear layers to align with the dimension of LLMs, therefore providing a more cohesive understanding of the video content when combined with the visual and acoustic modalities. By incorporating BEATs into VideoLLaMA, we address the challenge of synchronizing audio-visual data points. The encoder's ability to capture temporal dynamics aligns with the STC Connector employed in the visual branch, ensuring a seamless integration of audio-visual features.

\paragraph{Large Language Model Backbone}
Like its predecessor, VideoLLaMA 2 adopts the instruction-following large language models (LLMs) as its language decoder. We do not extensively search the optimal LLMs for VideoLLaMA 2 but use Mistral-Instruct~\citep{jiang2023mistral}, Mixtral-Instruct~\citep{jiang2024mixtral} and Qwen2-Instruct~\citep{qwen2} across all of the experiments. We leave the exploration of other popular LLMs, such as Gemma-IT~\citep{team2024gemma} and LLaMA3-Instruct~\citep{dubey2024llama}, for future work. We also include the results of VideoLLaMA 2 with Qwen2-Instruct~\citep{qwen2} as language decoder in Section~\ref{sec:evaluation}. 

\begin{figure}[t]
\centering
\includegraphics[width=1.0\textwidth]{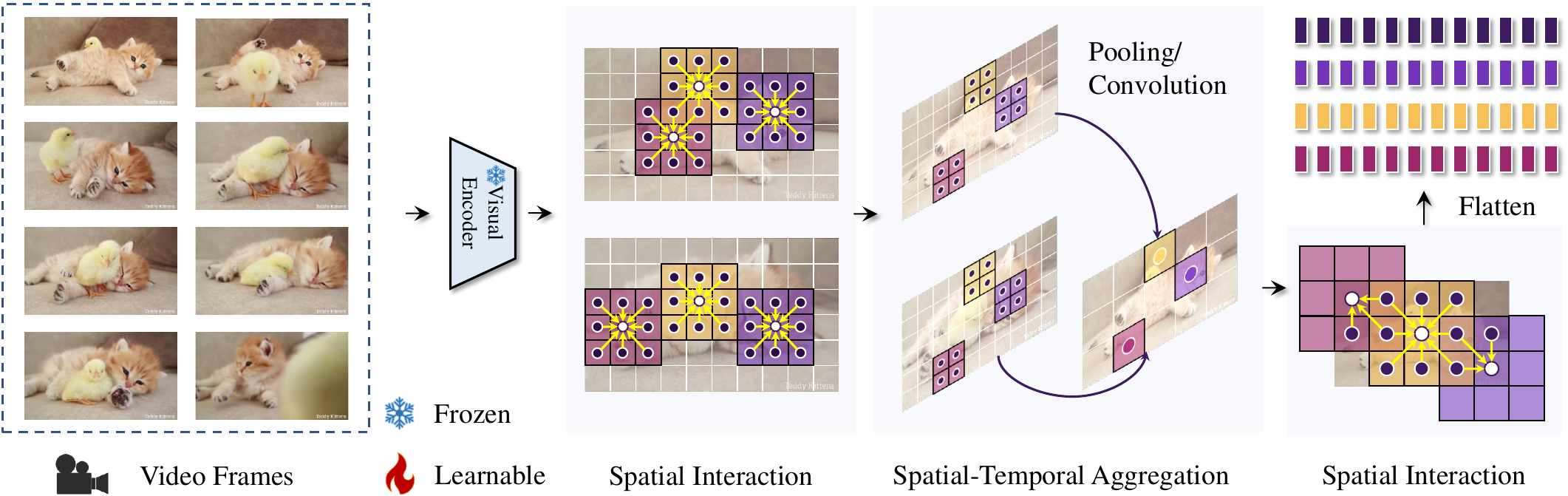} 
\caption{\textbf{The pipeline} of STC connector. The video frames are first encoded into features frame by frame, and then processed via our proposed STC connector~(two spatial interaction modules and one spatial-temporal aggregation module). We adopt RegStage to implement "Spatial interaction" and 3D convolution to implement "Spatial-Temporal Aggregation". 
}
\label{fig:stc_connector}
\end{figure}

\begin{table}[t]
\centering                         
\renewcommand{\arraystretch}{1.4}  
\setlength{\tabcolsep}{0.5mm}      
\footnotesize                      
\resizebox{1\textwidth}{!}{
\begin{tabular}{x{35}x{80}x{37}|x{60}x{72}x{95}x{30}}
\thickhline
\multirow{3}{*}{\bf{RegStage}} & \multirow{3}{*}{\bf{Downsample}} & \multirow{3}{*}{\bf{\# Tokens}} & \bf{MV-Bench} & \bf{Egoschema} & \bf{ActivityNet-QA} & \multirow{3}{*}{\bf{Avg.}} \\
& & & \scriptsize{Video QA (Acc.)} & \scriptsize{Long Video QA (Acc.)} & \scriptsize{Open-ended Video QA (Acc.)} & \\
& & & \scriptsize{\citep{li2023mvbench}} & \scriptsize{\citep{mangalam2024egoschema}} & \scriptsize{\citep{yu2019activitynet}} & \\
\hline
       & 2D Pool\ \ \ \ (1, 2, 2) & 1152 & 46.9 & 36.5 & 49.8 & 44.4 \\
       & 2D Conv\ \ (1, 2, 2)     & 1152 & 43.9 & 32.6 & 48.6 & 41.7 \\
       & 3D Pool\ \ \ \ (8, 1, 1) & 576  & 43.7 & 38.7 & 47.5 & 43.3 \\
       & 3D Pool\ \ \ \ (2, 2, 2) & 576  & 44.5 & 41.1 & 48.1 & 44.6 \\
       & 3D Conv\ \ (2, 2, 2)     & 576  & 44.8 & 39.0 & 45.6 & 43.1 \\
\hdashline
\cmark & 2D Pool\ \ \ \ (1, 2, 2) & 1152 &  46.7    &   35.3   &     46.9   &  43.0 \\
\cmark & 2D Conv\ \ (1, 2, 2)     & 1152 & 45.3 & 33.5 & 48.0 & 42.3 \\
\cmark & 3D Pool\ \ \ \ (2, 2, 2) & 576  & 44.4 & 39.1 & 46.0 & 43.2 \\
\rowcolor{alicegreen}
\cmark & 3D Conv\ \ (2, 2, 2)     & 576  & 45.5 & 42.2 & 47.6 & \textbf{45.1} \\
\thickhline
\end{tabular}}
\vspace{3pt}
\caption{\textbf{Empirical Study} of STC Connector design choices: Spatial Interaction~(RegStage), Spatial-Temporal Aggregation~(Downsample). To accelerate such empirical exploration, we train all of the models on the dataset provided by Video-LLaVA~\citep{lin2023video}. The \textbf{\textcolor{alicedarkgreen}{green}} line represents our chosen STC connector design. Furthermore, we default to sample 8 frames in this study.
}
\label{tab:tc_abl}
\end{table}

\paragraph{Spatial-Temporal Convolution Connector} 
\label{sec.connector}
The overall architecture and the workflow of our STC connector are depicted in Figure~\ref{fig:stc_connector}.
We primarily follow three principles when designing the video-language connector: 1) Maintaining the spatial-temporal order in the output visual tokens; 2) Reducing the number of spatial-temporal tokens; 3) Alleviating information loss during spatial-temporal downsampling.
\listnumber{\one} The first principle guides us to avoid using resampler architecture~\citep{li2023blip2bl,alayrac2022flamingoav} because the resampling operation does not guarantee the preservation of spatial-temporal order. 
This may cause suboptimal convergence because autoregressive models~(i.e., LLM backbone) are highly dependent on consistent token order between training and inference. Therefore, we only consider the operations of convolution or pooling when building our connector. 
\listnumber{\two} According to the second principle, we insert 3D downsample operator to compress spatial-temporal tokens.
\listnumber{\three} To complement the information loss caused by the spatial-temporal downsampling, we insert RegStage~(a strong convolution block~\citep{radosavovic2020designing}) before and after spatial-temporal downsampling. To empirically investigate the effectiveness of the designs above, we establish a quick but reasonable architectural search using the training data from Video-LLaVA~\citep{lin2023video}. We select three representative video understanding benchmarks: Egoschema~\citep{mangalam2024egoschema}, MV-Bench~\citep{li2023mvbench} and ActivityNet-QA~\citep{yu2019activitynet} as the testbed\footnote{See Section~\ref{video_understanding} for benchmark details and evaluation protocols.} and the comparison results are listed in Table~\ref{tab:tc_abl}. As can be seen, 3D convolution together with the ReStage block (i.e., the \textbf{\textcolor{alicedarkgreen}{green}} line), which forms our STC connector, works the best in terms of the average performance. Another interesting finding is that almost all of the 3D downsampling designs perform better than the 2D ones on Egoschema, suggesting that the early fusion of frame-level features is beneficial for long video understanding.

\begin{algorithm}[t]
\caption{PyTorch pseudo-code for STC connector}
\label{alg:stc}
\begin{lstlisting}[language=Python]
import torch.nn as nn
from timm.models.regnet import RegStage

class STCConnector(nn.Module):
    def __init__(self, config, depth, mlp_depth):
        # Temporal and spatial downsampling factor
        td, sd = config.td, config.sd
        # Input and output hidden dimension
        in_size, out_size = config.in_size, config.out_size
        # The first RegStage block
        self.s1 = RegStage(
            depth=depth, in_chs=in_size, out_chs=out_size)
        # Conv3D downsampler
        self.downsampler = nn.Conv3d(
                in_channels=out_size,
                out_channels=out_size,
                kernel_size=(td, sd, sd))
        # The second RegStage block
        self.s2 = RegStage(
            depth=depth, in_chs=out_size, out_chs=out_size)
        self.proj = build_mlp(mlp_depth, out_size, out_size)
        
    def forward(self, x):
        x = self.s1(x)
        x = self.downsampler(x)
        x = self.s2(x)
        x = self.proj(x)
        return x

\end{lstlisting}
\end{algorithm}

\section{Training}
In this section, we detail the training process for the Video-Language and Audio-language branches, followed by the joint training that integrates both modalities.

\begin{table}[t]
\centering                         
\renewcommand{\arraystretch}{1.4}  
\setlength{\tabcolsep}{1.5mm}      
\footnotesize                      
\begin{tabular}{y{70}|x{140}|x{48}x{48}x{48}}
\thickhline
\textbf{Modality} & \multicolumn{1}{c|}{\textbf{Dataset}} & \textbf{Original} & \textbf{Used} & \textbf{Ratio\%} \\ 
\hline
 & Panda-70M~\citep{chen2024panda} & 70M & 2.8M & 4\% \\
 & WebVid-10M~\citep{Bain21}       & 10M & 4M & 40\% \\
 & VIDAL-10M~\citep{zhu2023languagebind} & 10M  & 2.8M & 28\% \\
\multirow{-4}{*}{Video-Text} & InternVid-10M~\citep{wang2023internvid} & 10M & 650K & 6.5\% \\ 
\hdashline
\rowcolor{gray!10}
 & CC-3M~\citep{changpinyo2021conceptual} & 3M & 595K & 19.8\% \\
\rowcolor{gray!10}
\multirow{-2}{*}{Image-Text} & DCI~\citep{urbanek2023picture} & 7.8K & 7.8K & 100\%                     \\ 
\hline
Vision-Language & Total & 103M & 12.2M & 11.8\% \\
\thickhline
\end{tabular}
\vspace{3mm}
\caption{Video-Language Pre-training Data Statistics. We directly adopt the filtered version of CC-3M from~\cite{liu2023visualit}.}
\label{tab:vl_pretrain}
\end{table}

\begin{table}[t]
\centering                         
\renewcommand{\arraystretch}{1.4}  
\setlength{\tabcolsep}{1.5mm}      
\footnotesize                      
\begin{tabular}{y{70}|x{55}|x{50}y{190}}
\thickhline
\textbf{Modality} & \multicolumn{1}{c|}{\textbf{Task}} & \textbf{\# Samples} & \textbf{Dataset} \\ 
\hline
 & Captioning     & 23K & VideoChat, {\color{gray} In-house data} \\
 & \cellcolor{gray!10} Classification & \cellcolor{gray!10} 79K & \cellcolor{gray!10} Kinetics-710, SthSthv2 \\
 & VQA             & 161K & \begin{tabular}[c]{@{}l@{}}NExTQA, CLEVRER, EgoQA, Tgif, WebVidQA,\\  RealworldQA, Hm3d\end{tabular} \\
\multirow{-4}{*}{Video-Text} & \cellcolor{gray!10} Instruction & \cellcolor{gray!10} 225K & \cellcolor{gray!10} \begin{tabular}[c]{@{}l@{}}Valley, VideoChatGPT, VideoChat, VTimeLLM,\\  VideoChat2 \end{tabular} \\ 
\hdashline
 & Captioning     & 82K & ShareGPT4V \\
 & \cellcolor{gray!10} VQA & \cellcolor{gray!10} 198K & \cellcolor{gray!10} LLaVA \\
\multirow{-3}{*}{Image-Text} & Instruction & 466K & LLaVA \\
\hdashline
\multirow{-1}{*}{Pure Text} & \cellcolor{gray!10}Instruction & \cellcolor{gray!10}120K & \cellcolor{gray!10}Magpie, ALLaVA \\
\thickhline
\end{tabular}

\vspace{3mm}
\caption{Video-Language Multi-task Fine-tuning Data Statistics.}
\label{tab:vl_finetune}
\end{table}

\subsection{Video-Language Training}
\subsubsection{Pre-training}
For the pre-training stage, we utilize a large-scale, weakly labeled, web-crawled dataset of image-text and video-text pairs, sourced from several publicly accessible databases. The video-text sources include Panda-70M~\citep{chen2024panda}, VIDAL-10M~\citep{zhu2023languagebind}, WebVid-10M~\citep{Bain21}, and InternVid-10M~\citep{wang2023internvid}, while the image-text sources include CC-3M~\citep{changpinyo2021conceptual} and DCI~\citep{urbanek2023picture}.

During this stage, the vision encoder and the large language model are frozen, and only the connector is optimized. The input video frames are evenly sampled and resized to $336 \times 336$ pixels. The training objective is to minimize the cross-entropy loss of the text tokens. 

\subsubsection{Multi-task Fine-tuning}
\label{video_llm_instruct}
In the second stage of multi-task fine-tuning, we incorporate high-quality, fine-grained multi-modal annotations using both video-text and image-text data. As summarized in Table~\ref{tab:vl_finetune}
, VideoLLaMA 2 is fine-tuned on four tasks simultaneously. For video captioning, we use training samples from VideoChat~\citep{li2023videochat} and in-house collected data. For video classification and VQA tasks, we utilize a mixture of publicly available datasets, including Kinetics-710~\citep{kay2017kinetics}, SthSthv2~\citep{goyal2017something}, NExtQA ~\citep{xiao2021next}, CLEVRER~\citep{yi2019clevrer}, EgoQA ~\citep{fan2019egovqa}, Tgif~\citep{li2016tgif}, WebVidQA ~\citep{yang2021just}, RealworldQA ~\citep{xai2024grok15v}, and Hm3d ~\citep{ramakrishnan2021habitat}. To enhance instruction-following capabilities, we collect and integrate data from multiple frontier Video-LLMs, such as Valley ~\citep{luo2023valley}, VideoChatGPT ~\citep{maaz2023videochatgpt}, VideoChat ~\citep{li2023videochat}, VTimeLLM ~\citep{huang2023vtimellm}, and VideoChat2 ~\citep{li2023mvbench} to improve generalization. Additionally, we mix image captioning data from ShareGPT4V ~\citep{chen2023sharegpt4v} and image VQA and instruction-following data from LLaVA ~\citep{liu2023visualit,liu2023improved} to maintain the capabilities in understanding static visual concepts. To further improve the instruction-following abilities, we incorporate pure text data from Magpie~\citep{xu2024magpie} and ALLaVA~\citep{xu2023wizardlm,chen2024allava}.

In this stage, the visual encoder is frozen, and we optimize the language model and spatial-temporal connector. The training objective remains consistent with the pre-training stage. 

\begin{table}[t]
\centering
\renewcommand{\arraystretch}{1.4}  
\setlength{\tabcolsep}{1.5mm}      
\footnotesize
\begin{tabular}{y{130}|x{50}y{190}}
\thickhline
\textbf{Multi-stage} & \textbf{\# Samples} & \textbf{Data Sources} \\
\hline
Pre-training       & 400K & WavCaps  \\
\hdashline
\rowcolor{gray!10}
Instruction Tuning & 698K & \begin{tabular}[c]{@{}l@{}}ClothoAQA, WavCaps, AudioCaps, Clotho, \\ 
MusicCaps, VGGSound, UrbanSound8K, \\
ESC50, TUT2017, TUT2016, VocalSound \\
\end{tabular} \\
\hdashline
Audio-Video Joint Training & 836K & \begin{tabular}[c]{@{}l@{}}AVQA, AVQA-music, AVSD, VGGSound, \\
AVinstruct, MusicCaps, AudioCaps, VocalSound, \\
WavCaps, UrbanSound8K, Clotho, ClothoAQA, \\
TUT2017, TUT2016, Evol-Instruct, SthSthv2, \\
LLaVA, Kinetics-710, VideoChat, NExTQA, \\
Valley, EgoQA, CLEVRER, Tgif, ShareGPT4V, \\
{\color{gray} In-house data} \\  
\end{tabular} \\
\hline
Total& 1.9M & - \\
\thickhline
\end{tabular}
\vspace{3mm}
\caption{Datasets used in multi-stage audio-language training.}
\label{tab:datasets}
\end{table}

\subsection{Audio-Language Training}
Audio-language training of our VideoLLaMA 2 starts with the language decoder being initialized as the language model of the fine-tuned video model in Sec.~\ref{video_llm_instruct}. Similar to video-language training, audio-language training undergoes audio-language pre-training and multi-task fine-tuning.
\subsubsection{Pre-training}
In the initial phase, we focus on the foundational aspect of audio understanding by leveraging WavCaps \citep{mei2023wavcaps}, a comprehensive dataset comprising approximately 400,000 audio samples. Each sample is meticulously annotated for audio captioning tasks, aimed at training models to generate descriptive text based on audio inputs. This dataset serves as a crucial base for our models to learn intricate patterns in audio data, thereby preparing them for more complex audio-language tasks.

In this stage, the audio encoder and the large language model (LLM) are frozen, concentrating optimization exclusively on the audio projector. The primary training objective is to minimize the next token prediction loss over the textual responses, enhancing the model's capability to understand and map audio data to textual representations. This approach ensures that the audio processing components effectively leverage the optimized language model to achieve more precise audio-text alignment.

\subsubsection{Multi-task Fine-tuning}
The second stage of audio-language training aims to enhance the versatility and applicability of our model through multi-task learning, involving a variety of datasets tailored to different audio processing tasks. ClothoAQA \citep{lipping2022clotho}, with about 1,500 entries, is utilized for refining our model's capabilities in question answering based on audio cues, with each sample enriched by six associated questions sourced through crowdsourcing. The instruction tuning phase also capitalizes on the continued use of WavCaps \citep{mei2023wavcaps}, alongside AudioCaps \citep{kim2019audiocaps} and Clotho \citep{drossos2020clotho}, which together contribute about 453,000 audio samples for audio-text training. MusicCaps \citep{agostinelli2023musiclm}, comprising roughly 5,400 entries, extends our model's exposure to the musical domain, specializing in music-based captioning. For the purpose of sound event classification, we engage with VGGSound \citep{chen2020vggsound}, which offers over 190,000 labeled audio clips of diverse acoustic events. UrbanSound8k \citep{10.1145/2647868.2655045} and ESC50 \citep{piczak2015esc} provide additional layers of complexity with 8000+ and 2,000 urban and miscellaneous environmental sounds, respectively. TUT2017 \citep{mesaros2017dcase}, TUT2016 \citep{Mesaros2016_EUSIPCO}, and VocalSound \citep{gong2022vocalsound} enrich our dataset collection with more than 4,000, 1100, and 15,000 samples respectively, focusing on general sound events and human vocal sounds classification, thus broadening our model's acoustic perceptual skills.

At this stage, the audio encoder and audio projector are optimized, with the LLM remaining frozen, maintaining a consistent training objective to minimize text label cross-entropy loss as established in the pre-training stage.

\subsection{Audio-Video Joint Trainng}
The third phase shifts our focus towards the integration of audio and visual modalities, aiming to harness and understand the interactions between these two crucial aspects of multimodal perception. This phase incorporates audio-visual datasets including AVQA \citep{yang2022avqa} and AVQA-music \citep{li2022learning}, featuring around 57,000 and 35,000 entries, respectively. These datasets are specifically designed for audio-visual question answering, challenging our model to not only perceive but also interpret cross-modal content effectively. The AVSD \citep{alamri2019audio} dataset, consisting of approximately 80,000 pairs, is pivotal for developing and refining audio-visual dialogue systems. Furthermore, VGGSound \citep{chen2020vggsound} reappears with a substantial subset dedicated to audio-visual classification tasks, reinforcing the dataset's utility and versatility in our multimodal training framework. Additionally, we incorporate the AVInstruct \citep{ye2024cat} dataset, an audio-visual instruction dataset emphasizing co-learning on dynamic audio-visual pairs to address diverse AVQA tasks.

In addition to these audio-visual datasets, we also leverage several video-only datasets that enrich our system's capacity to process and analyze diverse forms of input. For video-text pairs, our resources include SthSthv2~\citep{goyal2017something} with 20K entries, EgoQA~\citep{fan2019egovqa} featuring 7.8K, Tgif~\citep{li2016tgif} contributing 24.7K, Kinetics-710~\citep{kay2017kinetics} incorporating 19.5K, and VideoChat2~\citep{li2023mvbench} with 9.5K. Other video-text datasets such as CLEVRER (2K)~\citep{yi2019clevrer} and Valley (35.5K) ~\citep{luo2023valley} enhance our model's robustness in understanding video content in conjunction with textual cues. Additionally, NExTQA ~\citep{xiao2021next}, though smaller with 1K pairs, serves as a valuable supplement. To further enhance the image-text interaction, we incorporate image-text datasets such as ShareGPT4V ~\citep{chen2023sharegpt4v} with a substantial 82K entries and LLaVA ~\citep{liu2023visualit,liu2023improved} using 62K pairs, providing the necessary frameworks for visual comprehension alongside textual analysis. 

Alongside these, we incorporate a broad range of audio-text pair datasets to bolster our model’s multimodal capabilities. Specifically, we include AudioCaps ~\citep{kim2019audiocaps} with 49.3K entries, VocalSound ~\citep{gong2022vocalsound} offering 15K, and UrbanSound8k \citep{10.1145/2647868.2655045} with 8K, which supply extensive audio-text data for urban and vocal soundscapes. Additionally, we utilize Clotho ~\citep{drossos2020clotho} with 19.3K entries, Wavcaps ~\citep{mei2023wavcaps} with 100K, MusicCaps ~\citep{agostinelli2023musiclm} providing 5.4K, and niche datasets such as TUT2017 ~\citep{mesaros2017dcase} and TUT2016 ~\citep{Mesaros2016_EUSIPCO} with 4.6K and 1.1K entries respectively, which focus on various environmental sounds. Furthermore, ClothoAQA ~\citep{lipping2022clotho} with 1.5K pairs adds to our resources for query-answering capabilities within audio-text contexts. These datasets collectively enable our model to engage with a rich variety of sound types and contexts, making it adept at handling complex audio-related tasks.

Finally, our text dataset, evol-instruct ~\citep{xu2023wizardlm}, consisting of 23.2K examples, rounds out our multimodal approach, facilitating a comprehensive training environment that ensures our models can effectively follow instructions and interpret a wide range of multimodal inputs.

In this stage, audio tracks are extracted from videos and cut to align the video clips. These audio clips are then truncated or padded to the same duration as the audio-language traning stage. For videos that lack an audio track, we fill the waveform with zeros to ensure uniformity across all data samples. During this training phase, we sample data for each batch in a 2:1:1 ratio of audio-visual data to visual data to audio data, ensuring that all audio-visual data is covered in one epoch. The video encoder remains frozen while we optimize the audio/video projector and the audio encoder, alongside the unfrozen LLM. The training objectives align with the instruction tuning stage, ensuring a coherent and effective progression in our multimodal training approach. By effectively leveraging the synchronized audio-visual data, VideoLLaMA 2 achieves a deeper understanding of multimodal content, thereby enhancing its performance across a spectrum of multimedia analysis tasks.

\section{Implementation}
Our VideoLLaMA2 is built upon LLaVA 1.5 library~\citep{liu2023improved}\footnote{\url{https://github.com/haotian-liu/LLaVA}}. Across all VideoLLaMA 2 variants, we use \texttt{clip-large-336}\footnote{\url{https://huggingface.co/openai/clip-vit-large-patch14-336}} as the primary visual encoder, though \texttt{siglip-so400m-384}\footnote{\url{https://huggingface.co/google/siglip-so400m-patch14-384}} is preferred in later variants due to its superior performance. For audio encoding, we utilize \texttt{Fine-tuned BEATs\_iter3+(AS2M)(cpt2)}\footnote{\url{https://1drv.ms/u/s!AqeByhGUtINrgcpj8ujXH1YUtxooEg?e=E9Ncea}}. The language decoders are initialized with either \texttt{Mistral-7B-Instruct}\footnote{\url{https://huggingface.co/mistralai/Mistral-7B-Instruct-v0.2}}, \texttt{Mixtral-8x7B-Instruct}\footnote{\url{https://huggingface.co/mistralai/Mixtral-8x7B-Instruct-v0.1}}, or, in some model variants, \texttt{Qwen2-7B-Instruct}\footnote{\url{https://huggingface.co/Qwen/Qwen2-7B-Instruct}} and \texttt{Qwen2-72B-Instruct}\footnote{\url{https://huggingface.co/Qwen/Qwen2-72B-Instruct}}. The audio branch training utilizes the video LLM initialized using the weights of the VideoLLaMA 2 (7B) model, which is trained on the two-stage video branch with 16 frames. We do not do any hyperparameter tuning during both pre-training and fine-tuning. Instead, we empirically set the global batch size and the learning rate as 1,024 and 1e-3 for pre-training and 2,048 and 2e-5 for fine-tuning. For the video-only training stage, VideoLLaMA 2 is pre-trained for just one epoch, followed by a fine-tuning process lasting up to three epochs. In the audio-only training, we also pre-train VideoLLaMA 2 for a single epoch, but limit fine-tuning to two epochs. For the audio-visual joint training stage, the pre-trained model undergoes fine-tuning for up to two epochs.

\section{Model Evaluation}
\label{sec:evaluation}
In this section, we present a comprehensive evaluation of VideoLLaMA 2, comparing it with other frontier models on various video and audio understanding benchmarks. The evaluation includes both quantitative metrics and qualitative analyses, highlighting the strengths and advancements of our model in handling complex multimodal data.

\subsection{Video Understanding}
\label{video_understanding}
\subsubsection{Evaluation Benchmarks}
We conduct extensive evaluations on Multi-choice Video Question Answering (MC-VQA), Open-Ended Video Question Answering (OE-VQA), and Video Captioning (VC) tasks to systematically assess the video understanding capabilities of VideoLLaMA 2.

\paragraph{MC-VQA} For the MC-VQA task, we select EgoSchema~\citep{mangalam2024egoschema}, Perception-Test~\citep{patraucean2024perception}, MV-Bench~\citep{li2023mvbench}, and VideoMME~\citep{fu2024video}. We report the top-1 accuracies for all benchmarks. For VideoMME, due to some unknown issues when getting the subtitles, we only report the results under the setting of ``w/o subs''. 

\paragraph{OE-VQA} For open-ended question answering, we conduct comparative studies using the MSVD-QA~\citep{xu2016msr}, ActivityNet-QA~\citep{yu2019activitynet}, and Video-ChatGPT~\cite{maaz2023videochatgpt} benchmarks. Following the protocols of \cite{maaz2023videochatgpt}, we employ a GPT-assisted evaluation to assess the quality of the generated answers. Specifically, GPT-3.5 provides a binary "Yes-or-No" decision on the correctness of answers, and we report the percentage of "Yes" responses as Accuracy.

\paragraph{VC} For the video captioning task, we perform experiments on the newly introduced Multi-Source Video Caption (MSVC) benchmark. MSVC is introduced to address limitations in existing video caption benchmarks, MSVC samples 500 videos with human-annotated captions from MSVD~\citep{chen2011collecting}, MSRVTT~\citep{xu2016msr}, and VATEX~\citep{wang2019vatex}, ensuring diverse scenarios and domains. Traditional evaluation metrics rely on exact match statistics between generated and ground truth captions, which are limited in capturing the richness of video content. Thus, we use a ChatGPT-assisted evaluation similar to \cite{maaz2023videochatgpt}. Both generated and human-annotated captions\footnote{Each video in the MSVC benchmark is annotated with multiple human-written captions, covering different aspects of the video. This comprehensive annotation ensures a robust and thorough evaluation of Video-LLMs.} are evaluated by GPT-3.5-turbo (0613) for Correctness of Information and Detailed Orientation~\footnote{We also include evaluation results on several other popular video benchmarks~\citep{MLVU,li2024videovista,fang2024mmbenchvideo,wang2024tarsierrecipestrainingevaluating} in Appendix~\ref{app:additional results}.}.

\begin{table}[t]
\centering                         
\renewcommand{\arraystretch}{1.4}  
\setlength{\tabcolsep}{0.5mm}      
\scriptsize                        
\begin{tabular}{y{100}x{30}|x{40}x{50}x{35}x{40}|x{40}x{40}}
\thickhline
\multirow{3}{*}{\bf{Model}} & \multirow{3}{*}{\bf{\# Frames}} & \multicolumn{4}{c|}{\bf{MC-VQA}} & \multicolumn{2}{c}{\bf{VC}} \\
\cline{3-8}
& &
\textbf{EgoSchema} &
\textbf{Perception-Test} &
\textbf{MVBench} & \textbf{VideoMME} & 
\multicolumn{2}{c}{\textbf{MSVC~(Score)}} \\
& & (Acc.) & (Acc.) & (Acc.) & (Acc.) & correctness & detailedness \\
\hline
\rowcolor{gray!20}
\multicolumn{8}{c}{\it{Proprietary Models}}\\
\hline
Gemini 1.0 Pro~\citep{team2023gemini}   & - & 55.7\official & 51.1\official & -    & -                       & -    & -  \\
Gemini 1.0 Ultra~\citep{team2023gemini} & - & 61.5\official & 54.7\official & -    & -                       & -    & -  \\
Gemini 1.5 Flash~\citep{team2024gemini} & - & -                   & -                   & -    & 70.3/75.0\leaderboard                       & 3.46\reproduce & 3.24\reproduce \\
Gemini 1.5 Pro~\citep{team2024gemini}   & - & 63.2\official & -                   & -    & \textbf{75.0}/\textbf{81.3}\leaderboard   & \textbf{3.67}\reproduce & \textbf{3.52}\reproduce \\
GPT4-V~\citep{openai2023gpt4v}          & - & 55.6\official & -    & 43.7\leaderboard & 59.9/63.3\leaderboard & 2.70\reproduce & 2.76\reproduce \\
GPT4-O~\citep{openai2024gpt4o}         & - & \bf{72.2}\official&- & -    & 71.9/77.2\leaderboard & -  & -  \\
Reka-Flash~\citep{ormazabal2024reka}    & - & -    & 56.4\official & -    & - & - & - \\
Reka-Core~\citep{ormazabal2024reka}     & - & -    & \bf{59.3}\official &-& -    & 2.61\reproduce & 2.73\reproduce \\
\hline
\rowcolor{gray!20}
\multicolumn{8}{c}{\it{Open-source Models}}\\
\hline
LLaMA-VID~(7B)        & 1 fps & 38.5\reproduce & 44.6\reproduce & 41.9\reproduce & 25.9/-\reproduce & 1.84\reproduce & 2.11\reproduce \\
Video-LLaVA~(7B)      & 8  & 38.4\reproduce & 44.3\reproduce & 41.0\reproduce & 39.9/41.6\leaderboard & 1.85\reproduce & 2.05\reproduce \\
VideoChat2~(7B)       & 16 & 42.2\reproduce & 47.3\reproduce & 51.1\official  & 33.7/-\leaderboard & 2.01\reproduce & 2.10\reproduce \\
LLaVA-NeXT-Video~(7B) & 32 & 43.9\reproduce & 48.8\reproduce & 46.5\reproduce & - & 2.40\reproduce & 2.52\reproduce \\ 
LLaVA-NeXT-Video~(32B) & 32 & 60.9\official & - & - & 60.2/63.0\official & - & - \\
PLLaVA~(34B) & 16 & - & - & 58.1\official & - & - & - \\
VILA 1.5~(34B) & - & 58.0\reproduce & - & - & 62.3/64.1\leaderboard & - & - \\
LLaVA-OneVision~(72B) & 32 & 62.0\official & - & 59.5\official & \textbf{66.3}/\textbf{69.6}\official & - & - \\
\hdashline
\rowcolor{alicegreen}
VideoLLaMA2~(7B)     & 8  & 50.5 & 49.6 & \multicolumn{1}{c}{53.4} & 
45.1/46.6 & 2.57 & 2.61 \\
\rowcolor{alicegreen}
VideoLLaMA2 (7B)     & 16  & 51.7 & 51.4 & \multicolumn{1}{c}{54.6} & 47.9/50.3 & 2.53 & 2.59 \\
\rowcolor{alicegreen}
VideoLLaMA2 (8x7B)   & 8  & 53.3 & 52.2 & \multicolumn{1}{c}{53.9} & 47.9/49.7 & 2.53 & 2.56 \\
\rowcolor{alicegreen}
VideoLLaMA2 (72B)   & 16  & \textbf{63.9} & \textbf{57.5} & \multicolumn{1}{c}{\textbf{62.0}} & 61.4/63.1 & 2.71 & 2.67 \\
\rowcolor{alicegreen}
VideoLLaMA2.1 (7B)  & 16  & 53.1 & 54.9 & 57.3 & 54.9/56.4 & \textbf{2.87} & \textbf{2.81} \\
\thickhline
\end{tabular}
\vspace{3pt}
\caption{\textbf{Main Results} on Multiple-Choice Video QA~(MC-VQA) and Video Captioning~(VC). We do not follow the latest version of Gemini 1.5~\citep{team2024gemini} to invoke explicit CoT on EgoSchema, which could give large performance gains.
{\textcolor{plum}{$\varheartsuit$}}: officially reported results. {$\diamondsuit$}: results retrieved from the leaderboard. {\textcolor{darkpurple}{$\spadesuit$}}: results reproduced by ourselves.}
\label{tab:benchmark_mcvqa_vc}
\end{table}

\subsubsection{Baselines}
To comprehensively evaluate the performance of VideoLLaMA 2, we compare it against a diverse set of baselines, including both proprietary and open-source frontier models. The included models are listed below:

\paragraph{Proprietary Models} The proprietary models selected for comparison are state-of-the-art multi-modal systems developed by leading companies. These models include Gemini 1.0 Series~\citep{team2023gemini}, Gemini 1.5 Series~\citep{team2024gemini}, GPT4-V~\citep{openai2023gpt4v}, GPT4-O~\citep{openai2024gpt4o}, Reka Series~\citep{ormazabal2024reka}, and Pegasus-1~\citep{jung2024pegasus}. These models represent the cutting-edge proprietary multi-modal understanding technologies and serve as benchmarks to gauge the performance of VideoLLaMA 2.

\paragraph{Open-Source Models} We also include several prominent open-source models to provide a broader context for our evaluations. Specifically, we compare our VideoLLaMA 2 with VistaLLaMA~\citep{ma2023vistallama}, ChatUniVi~\citep{jin2023chatunivi}, LLaMA-VID~\citep{li2023llama}, Video-LLaVA~\citep{lin2023video}, VideoChat2~\citep{li2023mvbench}, LLaVA-NeXT-Video Series\footnote{For a fair comparison, we do not include the preference-optimized LLaVA-NeXT-Video as baseline. However, we still provide the full comparison results between VideoLLaMA 2 and LLaVA-NeXT-Video-DPO in Appendix~\ref{sec:comp_llava_next_video_dpo} for the reference of readers.}~\citep{liu2024llavanext}, VILA 1.5~\citep{lin2024vila}, PLLaVA~\citep{xu2024pllava}, and LLaVA-OneVision~\citep{li2024llavaonevision}. These open-source models are crucial for evaluating the performance of VideoLLaMA 2 within the context of accessible and reproducible research.

\subsubsection{Evaluation Protocol}
All experiments, including the reproduction of baseline models, are conducted in a zero-shot manner to objectively assess generalization capabilities. For model decoding strategies, we use greedy search for all benchmarks except MSVC, where we apply sampling with a temperature of 0.2 to enhance the diversity and detailedness of the generated captions. We follow the original setup of baseline models regarding the number of frames used for each input video, ensuring fair and consistent comparisons across all models.

\subsubsection{Main Results}
\paragraph{Results on MC-VQA and VC}
The overall performance on multiple-choice video question answering (MC-VQA) and video captioning (VC) tasks are summarized in Table~\ref{tab:benchmark_mcvqa_vc}. VideoLLaMA 2 demonstrates strong performance compared to open-source models and shows competitive results against proprietary models in certain benchmarks.

For MC-VQA, VideoLLaMA 2 exhibits substantial improvements over open-source models. On the EgoSchema benchmark, VideoLLaMA 2-7B achieves an accuracy of $51.7\%$, outperforming the previous SOTA LLaVA-NeXT-Video ($43.9\%$) by a large margin. Similarly, on the Perception-Test and MV-Bench datasets, VideoLLaMA 2-7B attains accuracies of $51.4\%$ and $53.9\%$, respectively, surpassing other open-source models. Notably, VideoLLaMA 2 also outperforms the proprietary model GPT4-V ($43.7\%$) on the MV-Bench dataset.
Additionally, VideoLLaMA 2 shows competitive performance on the VideoMME benchmark with an accuracy of $48.4\%$, highlighting its robust capabilities in video understanding tasks.
Furthermore, scaling up the LLM backbone from Mistral (7B) to Mixtral (8$\times$7B) further enhances model performance in MC-VQA. This upscaling results in notable improvements across multiple benchmarks, with VideoLLaMA 2-8$\times$7B achieving the highest accuracies on Egoschema, Perception-Test, and Video-MME. 

In the VC task, VideoLLaMA 2 performs well on the MSVC benchmark, scoring $2.57$ in correctness and $2.61$ in detailedness. While these scores are slightly lower than GPT4-V's $2.70$ and $2.76$, they are higher than all other open-source models, demonstrating the model's strong capabilities in interpreting dynamic video content. 

\begin{table}[t]
\centering                         
\renewcommand{\arraystretch}{1.4}  
\setlength{\tabcolsep}{0.5mm}      
\scriptsize                        
\begin{tabular}{y{100}x{30}|x{40}x{40}|x{32}x{32}x{32}x{32}x{35}}
\thickhline
\multirow{2}{*}{\textbf{Model}} & \multicolumn{1}{c|}{\multirow{2}{*}{\textbf{\# Frames}}} & \bf{MSVD} & \bf{ActivityNet} & \multicolumn{5}{c}{\textbf{Video-ChatGPT~(Score)}} \\
\cline{3-9} 
& & (Acc./Score) & (Acc./Score) & Correctness & Detail & Context & Temporal & Consistency \\ 
\hline
\rowcolor{gray!20}
\multicolumn{9}{c}{\it{Proprietary Models}}\\
\hline
Gemini 1.0 Pro     & - & - & 49.8/-\official & -    & -    & -    & -    & -    \\
Gemini 1.0 Ultra   & - & - & 52.2/-\official & -    & -    & -    & -    & -    \\
Gemini 1.5 Pro     & - & - & 56.7/-\official & -    & -    & -    & -    & -    \\
GPT4-V             & - & - & 59.5/-\official & \textbf{4.09} & \textbf{3.88} & \textbf{4.37} & \textbf{3.94} & \textbf{4.02} \\
GPT4-O             & - & - & \textbf{61.9}/-\official & -    & -    & -    & -    & -    \\
Pegasus-1 & - & - & 59.9/-\official & 3.79\official    & 3.76\official    & 4.29\official    & 3.34\official    & 4.03\official    \\
\hline
\rowcolor{gray!20}
\multicolumn{9}{c}{\it{Open-Source Models}}\\
\hline
VideoLLaMA (7B)       & 8  & 51.6/2.5 & 12.4/1.1 & 1.96 & 2.18 & 2.16 & 1.82 & 1.79 \\
Video-ChatGPT (7B)    & 8  & 64.9/3.3 & 35.2/2.7 & 2.50 & 2.57 & 2.69 & 2.16 & 2.20 \\
VideoChat (7B)        & 8  & 56.3/2.8 & 26.5/2.2 & 2.23 & 2.50 & 2.53 & 1.94 & 2.24 \\
Chat-UniVi (7B)       & 8  & 65.0/3.6\official & 46.1/3.3\official & 2.89 & 2.91 & 3.46 & 2.89 & 2.81 \\
LLaMA-VID (7B)        & 1 fps  & 69.7/3.7\official & 47.4/3.3\official & 2.96 & 3.00 & 3.53 & 2.46 & 2.51 \\
Video-LLaVA (7B)      & 8  & 70.7/3.9\official & 45.3/3.3\official & 2.87 & 2.94 & 3.44 & 2.45 & 2.51 \\
VideoChat2 (7B)       & 16 & 70.0/3.9\official & 49.1/3.3\official & 3.02 & 2.88 & 3.51 & 2.66 & 2.81 \\
LLaVA-NeXT-Video (7B) & 32 & 67.8/3.5\reproduce & 53.5/3.2\official & 3.39\official & \textbf{3.29}\official & \textbf{3.92}\official	& 2.60\official & 3.12\official \\
LLaVA-NeXT-Video (32B) & 32 & - & 54.3/-\official & - & - & -	& - & - \\
PLLaVA (34B) & 16 & - & 60.9/-\official & \textbf{3.60}\official & 3.20\official & 3.90\official	& 2.67\official & 3.25\official \\
LLaVA-OneVision (72B) & 32 & - & \textbf{62.3}/-\official & - & - & -	& - & - \\
\hdashline
\rowcolor{alicegreen}
VideoLLaMA2 (7B)    & 8  & \textbf{71.7/3.9} & 49.9/3.3 & 3.09 & 3.09 & 3.68 & 2.63 & 3.25 \\
\rowcolor{alicegreen}
VideoLLaMA2 (7B)    & 16 & 70.9/3.8 & 50.2/3.3 & 3.16 & 3.08 & 3.69 & 2.56 & 3.14 \\
\rowcolor{alicegreen}
VideoLLaMA2 (8x7B)  & 8  & 70.5/3.8 & 50.3/3.4 & 3.08 & 3.11 & 3.64 & \textbf{2.67} & \textbf{3.26} \\
\rowcolor{alicegreen}
VideoLLaMA2 (72B)  & 8  & 71.0/3.8 & \textbf{55.2}/\textbf{3.4} & 3.23 & 3.11 & 3.71 & 2.62 & 3.12 \\
\rowcolor{alicegreen}
VideoLLaMA2.1 (7B)  & 16  & 70.6/3.8 & 53.0/3.4 & 3.30 & 3.18 & 3.78 & 2.66 & 3.20 \\
\thickhline
\end{tabular}
\vspace{3pt}
\caption{\textbf{Main Results} on Open-Ended Video QA~(OE-VQA) benchmarks.
{\textcolor{plum}{$\varheartsuit$}}: officially reported results. {\textcolor{darkpurple}{$\spadesuit$}}: results reproduced by ourselves. The numbers without marks are retrieved from~\cite{maaz2023videochatgpt,luo2023valley,liu2024llavanext}.
}
\label{tab:benchmark_oevqa}
\end{table}

\paragraph{Results on OE-VQA}
The performance on Open-Ended Video Question Answering~(OE-VQA) tasks is summarized in Table~\ref{tab:benchmark_oevqa}.
VideoLLaMA 2 demonstrates strong performance compared to both proprietary and open-source models across several benchmarks. For the MSVD dataset, VideoLLaMA 2-7B gets an accuracy of $71.7\%$ with a score of $3.9$, outperforming other open-source models, e.g., LLAVA-NeXT-Video~($67.8\%/3.5$) and VideoChat2~($70.0\%/3.9$). 

However, on the ActivityNet dataset, VideoLLaMA 2-7B attains an accuracy of $49.9\%$ with a score of $3.3$, which is slightly lower than LLAVA-NeXT-Video ($53.5\%/3.2$).
Similarly, on the Video-ChatGPT benchmark, VideoLLaMA 2-7B scores $3.09$ in correctness, $3.09$ in detail, $3.68$ in context, $2.63$ in temporal understanding, and $3.25$ in consistency. While VideoLLaMA 2 achieves high scores, it is outperformed by LLAVA-NeXT-Video in several metrics, particularly in correctness ($3.39$), detail ($3.29$), and context ($3.92$). These two benchmarks, both based on the ActivityNet~\citep{yu2019activitynet} dataset, typically include videos depicting a single human activity. Given that LLAVA-NeXT-Video~\citep{liu2024llavanext} is trained primarily on static image data with a comparably small amount of dynamic video data, it shows advantages in benchmarks where static visual information is crucial for answering questions. This suggests that training on massive static image data can be beneficial for video tasks that are heavily reliant on static visual information, a hypothesis we will explore further in future research.

\subsection{Audio Understanding}
To evaluate the audio understanding capabilities of VideoLLaMA 2, we conduct comprehensive evaluations on several established audio understanding benchmarks. These evaluations aim to measure the model's proficiency in interpreting and integrating audio information in conjunction with video data. We use ChatGPT 3.5 to assess the prediction of models and the prompt is provided in Appendix\ref{sec:gpt_eval}. The ChatGPT is required to provide the ``yes'' or ``no'' binary response, followed by a integer score to quantify the degree of match. 

\subsubsection{Evaluation Benchmarks}

We conduct extensive experiments on Audio-only Question Answering (AQA) task, followed by Open-Ended Audio-Video Question Answering tasks to assess the audio comprehension abilities of VideoLLaMA 2. 


\paragraph{AQA} For the AQA task, we select open-ended Clotho-AQA ~\citep{Samuel2022clothoaqa} and multiple-choice TUT2017 ~\citep{Annamaria2016} and VocalSound ~\citep{gong2022vocalsound} datasets as our benchmark and report accuracy as the evaluation metric.


\paragraph{OE-AVQA} For open-ended audio-video question answering, we adopt VGGSound ~\citep{chen2020vggsound} (15341 samples), AVSD~\citep{alamri2019audio} (18630 samples) and Music-AVQA~\citep{li2022avqamusic} (9129 samples) as benchmarks. For VGGSound and AVSD, we employ a GPT-assisted evaluation as the same as the protocols in OE-VQA. 

\subsubsection{Baselines}
To evaluate the performance of VideoLLaMA 2 within the realms of audio understanding and audio-visual integration, we compare it against an array of established and cutting-edge models in these fields.

\paragraph{Audio Understanding} To assess the audio understanding capabilities of VideoLLaMA 2-7B, we utilize Qwen-Audio ~\citep{Qwen-Audio} and Qwen2-Audio ~\citep{chu2024qwen2} as the benchmarks. Qwen-Audio and Qwen2-Audio are renowned for its robust performance across various audio understanding tasks, providing a strong comparison point for our model.

\paragraph{Audio-Visual Understanding} Expanding the evaluation to audio-visual integration, VideoLLaMA 2 is benchmarked against advanced models that specialize in handling multimodal data. This includes comparisons with PandaGPT \citep{pandagpt}, MacawLLM \citep{lyu2023macaw}, Video-LLaMA \citep{zhang2023video}, and X-InstructBLIP \citep{panagopoulou2023x}, which are adept at understanding complex multimodal scenes. Further, we examine newer integrative models like AV-LLM \citep{shu2023audio}, OneLLM \citep{han2024onellm}, CREMA \citep{yu2024crema}, and AVicuna \citep{tang2024avicuna}, which utilize high-quality audio-visual training datasets to enhance their multimodal understanding capabilities. 

\subsubsection{Main Results}
\begin{table}[t]
\centering                         
\renewcommand{\arraystretch}{1.4}  
\setlength{\tabcolsep}{1.5mm}      
\footnotesize
\begin{tabular}{y{110}x{60}|x{60}x{60}x{60}}
\thickhline
\textbf{Method} & \textbf{\# Hours} & \textbf{Clotho-AQA} & \textbf{TUT2017} & \textbf{VocalSound}\\
\hline
Qwen-Audio~(7B)      & ~137k & 57.90 & 64.90 & 92.89  \\
Qwen2-Audio~(7B)      & ~520k & - & - & \bf{93.92}  \\
\hdashline
\rowcolor{alicegreen}
VideoLLaMA2-AV~(7B)    & ~5k & 70.11 & \bf{78.40} & 93.19 \\
\rowcolor{alicegreen}
VideoLLaMA2.1-AV~(7B)    & ~5k & \bf{71.02} & 77.28 & 92.40 \\
\thickhline
\end{tabular}
\vspace{3pt}
\caption{Comparison with existing LLM-based methods on open-ended AQA (Clotho-AQA) and multiple-choice AQA (TUT2017 and VocalSound) benchmarks.}
\label{tab:audio_qa_comparison}
\end{table}

\begin{table}[t]
\centering                         
\renewcommand{\arraystretch}{1.4}  
\setlength{\tabcolsep}{1.5mm}      
\footnotesize
\begin{tabular}{y{110}x{60}|x{60}x{60}x{60}}
\thickhline
\textbf{Method} & \textbf{\# Pairs} & \textbf{MUSIC-QA} & \textbf{AVSD} & \textbf{VGGSound}\\
\hline
PandaGPT~(13B)      & 128M & 33.7 & 26.1 & 32.7 \\
Macaw-LLM~(7B)      & 0.3M & 31.8 & 34.3 & 36.1 \\
VideoLLaMA~(7B)     & 2.8M & 36.6 & 36.7 & 40.8 \\
X-InstructBLIP~(13B) & 32M & 44.5 & - & - \\
AV-LLM~(13B)        & 1.6M & 45.2 & 52.6 & 47.6 \\
OneLLM~(7B)        & 1007M & 47.6 & - & - \\
AVicuna~(7B)        & 1.1M & 49.6 & 53.1 & - \\
CREMA~(4B)        & - & 52.6(75.6) & - & - \\
\hdashline
\rowcolor{alicegreen}
VideoLLaMA2-AV~(7B)   & 1.9M & 79.2 & 57.2 & 70.9 \\
\rowcolor{alicegreen}
VideoLLaMA2.1-AV~(7B)   & 2.0M & \bf{80.9} & \bf{57.2} & \bf{71.4} \\
\thickhline
\end{tabular}
\vspace{3pt}
\caption{Comparison with existing LLM-based methods on Open-Ended Audio-Video QA~(OE-AVQA) benchmarks. \# Pairs: the adopted instruction-response pairs. Note: All baseline models for MUSIC-QA and AVSD are zero-shot. For the VGGSound dataset, the first three models are zero-shot, while the remaining are in-domain. The number in the bracket represents the enhanced result obtained through specialized model fine-tuning.}
\label{tab:video_avqa_comparison}
\end{table}
\paragraph{Results on AQA}
The experimental findings presented in Table~\ref{tab:audio_qa_comparison} delineate the compelling advantages of VideoLLaMA 2-AV (7B) and VideoLLaMA 2-AV (7B) on audio-only question answering (AQA). 

Using the widely recognized Qwen-Audio and Qwen2-Audio as benchmarks, both VideoLLaMA 2-A (7B) and VideoLLaMA 2-AV (7B) demonstrate exceptional performance across multiple datasets, showcasing remarkable efficiency and advanced learning capabilities despite being trained on significantly fewer hours of data. Specifically, VideoLLaMA 2-A (7B), with only 4k hours of audio data, outperforms Qwen-Audio (137k hours) across most benchmarks. It achieves 68.90\% on Clotho-AQA, a substantial improvement over Qwen-Audio's 57.90\%, and records 75.19\% on TUT2017, surpassing Qwen-Audio’s 64.90\% by 10.29\%. This highlights its robust understanding and interpretative capabilities in complex audio question-answering tasks. Meanwhile, in the VocalSound benchmark, where high accuracy is essential, VideoLLaMA 2-A (7B) achieves 92.73\%, closely matching Qwen2-Audio (7B)’s 93.92\% despite being trained on far fewer hours (4k vs. 520k).

Furthermore, VideoLLaMA 2-AV (7B), leveraging about 5k hours of multimodal (audio-visual) data, builds on the strengths of VideoLLaMA 2-A and achieves even higher performance. It records the highest scores on Clotho-AQA (70.11\%) and TUT2017 (78.40\%). On VocalSound, VideoLLaMA 2-AV (7B) reaches 93.19\%, further narrowing the gap with Qwen2-Audio (93.92\%). These results emphasize the effectiveness of the VideoLLaMA framework, showcasing both models' ability to achieve superior or comparable results with much fewer training hours. This highlights their efficiency and demonstrates the value of integrating audio-visual modalities for enhanced learning.

\paragraph{Results on OE-AVQA}
The performance results summarized in Table~\ref{tab:video_avqa_comparison} highlight VideoLLaMA 2-AV (7B)'s significant advancements compared to open-source models. This analysis underscores VideoLLaMA 2-AV (7B)'s capabilities across various AVQA benchmarks, emphasizing its superior performance.

In the MUSIC-QA benchmark, VideoLLaMA 2-AV (7B) demonstrates substantial competency with a score of 79.2\%. This score indicates a robust ability to analyze complex musical and audio cues, which is central to the MUSIC-QA benchmark. The model's performance, outreaching CREAM with specialized model fine-tuning at 75.6\%, underscores its advanced capabilities in handling intricate audio-visual interactions. In the AVSD benchmark, which focuses on audio-visual scene description involving dialogue understanding, VideoLLaMA 2-AV (7B) also demonstrates strong capabilities by scoring 57.2\%, illustrating its adeptness at integrating visual context with audio inputs to generate coherent scene descriptions. The VGGSound benchmark, which necessitates a deep understanding of both visual scenes and their auditory elements, sees VideoLLaMA 2-AV (7B) leading with a score of 70.9\%. This performance, much higher than AV-LLM’s 47.6\%, showcases VideoLLaMA 2-AV (7B)'s exceptional ability to interpret and synthesize information across both modalities effectively. It indicates a profound understanding of dynamic interactions and the contextual synthesis necessary in AVSSD (audio-visual sound source detection) tasks.

\section{Cases}

\begin{figure}[t]
\begin{subfigure}[b]{0.99\linewidth}
    \includegraphics[width=0.98\linewidth]{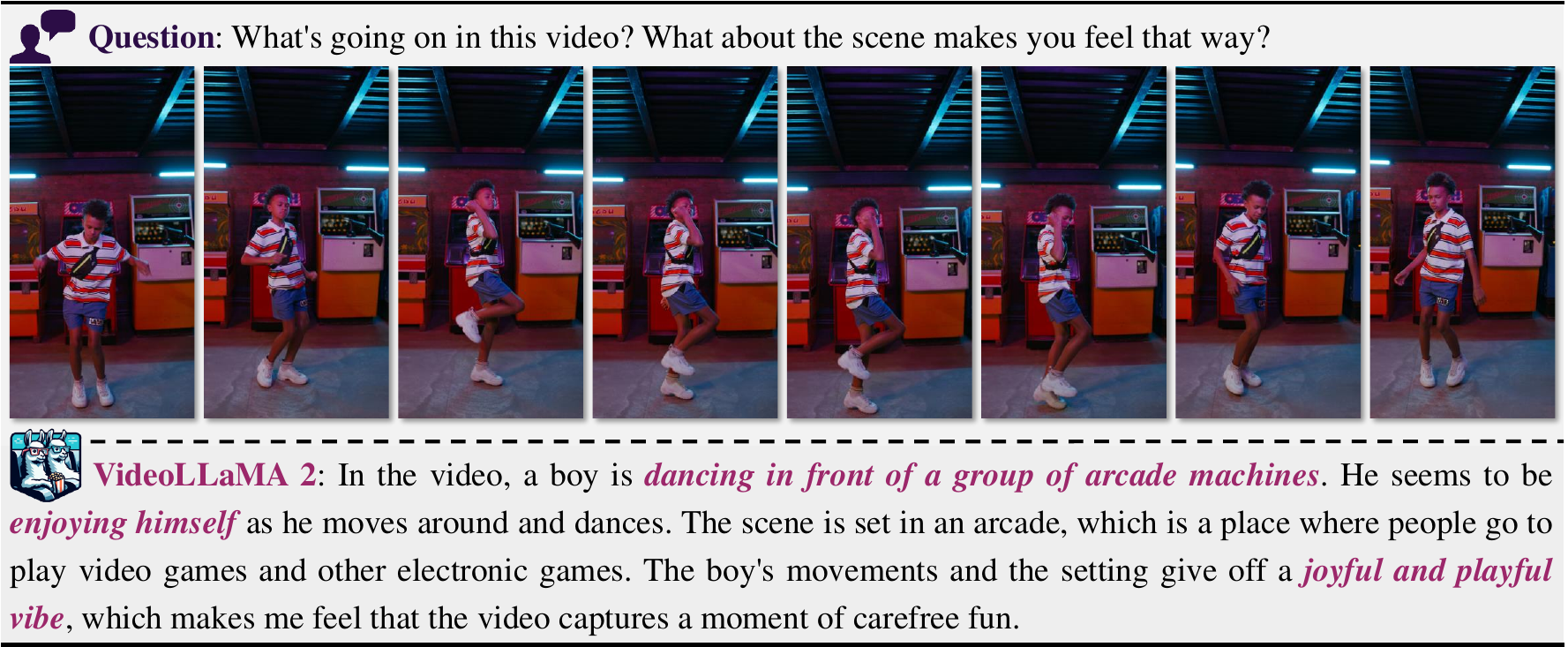}
    \caption{\textbf{Global Scene Description Case.}}
    \label{case:gsd}
\end{subfigure}
\hfill
\begin{subfigure}[b]{0.99\linewidth}
    \includegraphics[width=0.98\linewidth]{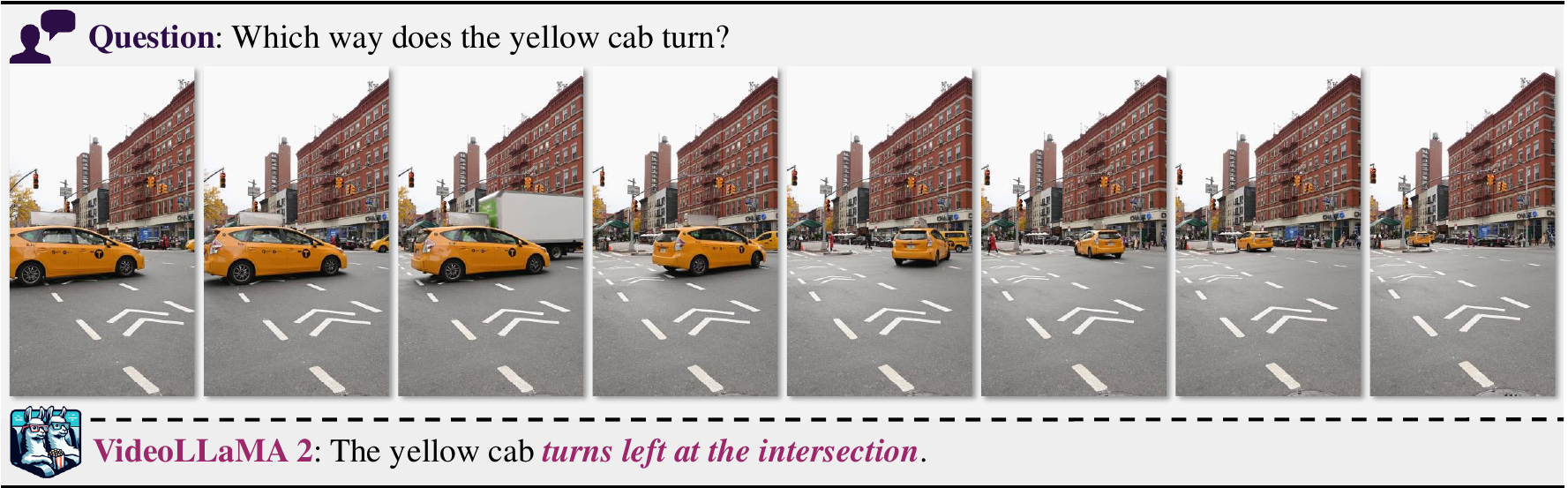}
    \caption{\textbf{Spatial-temporal Orientation Awareness Case.}}
    \label{case:stp}
\end{subfigure}
\hfill
\begin{subfigure}[b]{0.99\linewidth}
    \includegraphics[width=0.98\linewidth]{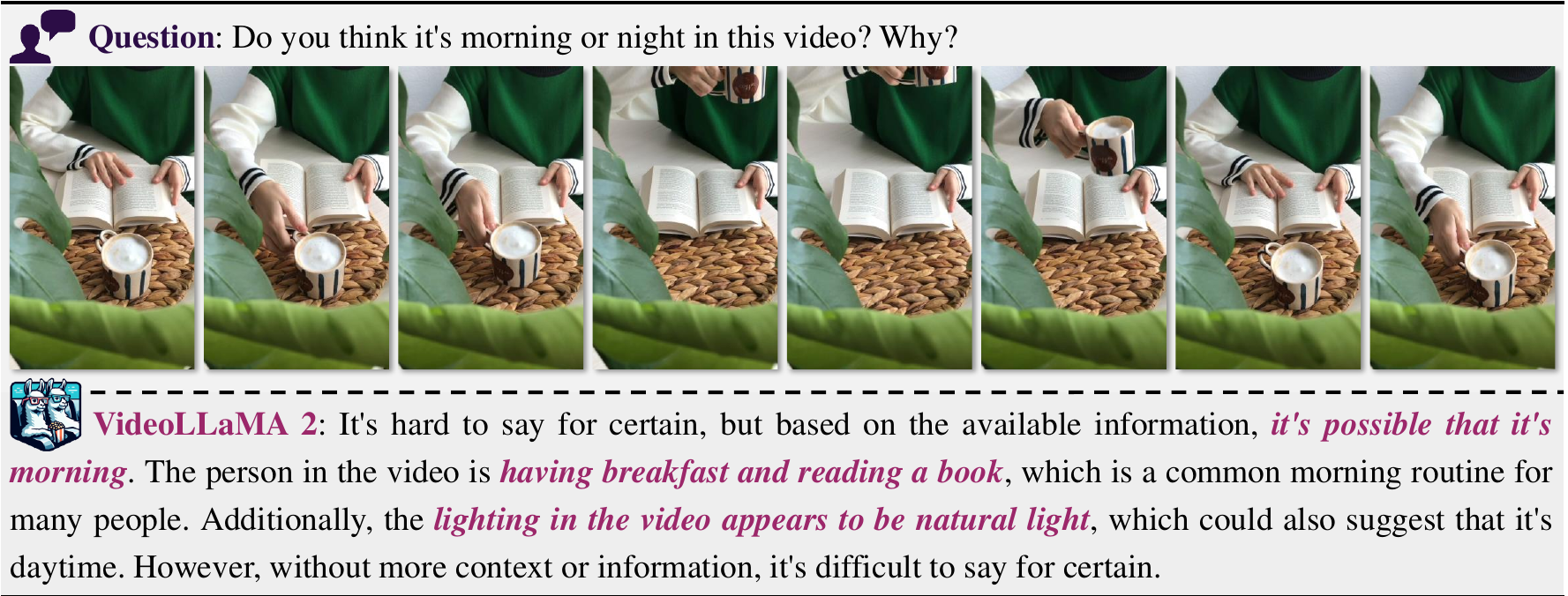}
    \caption{\textbf{Commonsense Reasoning Case.}}
    \label{case:csr}
\end{subfigure}
\hfill
\begin{subfigure}[b]{0.99\linewidth}
    \includegraphics[width=0.98\linewidth]{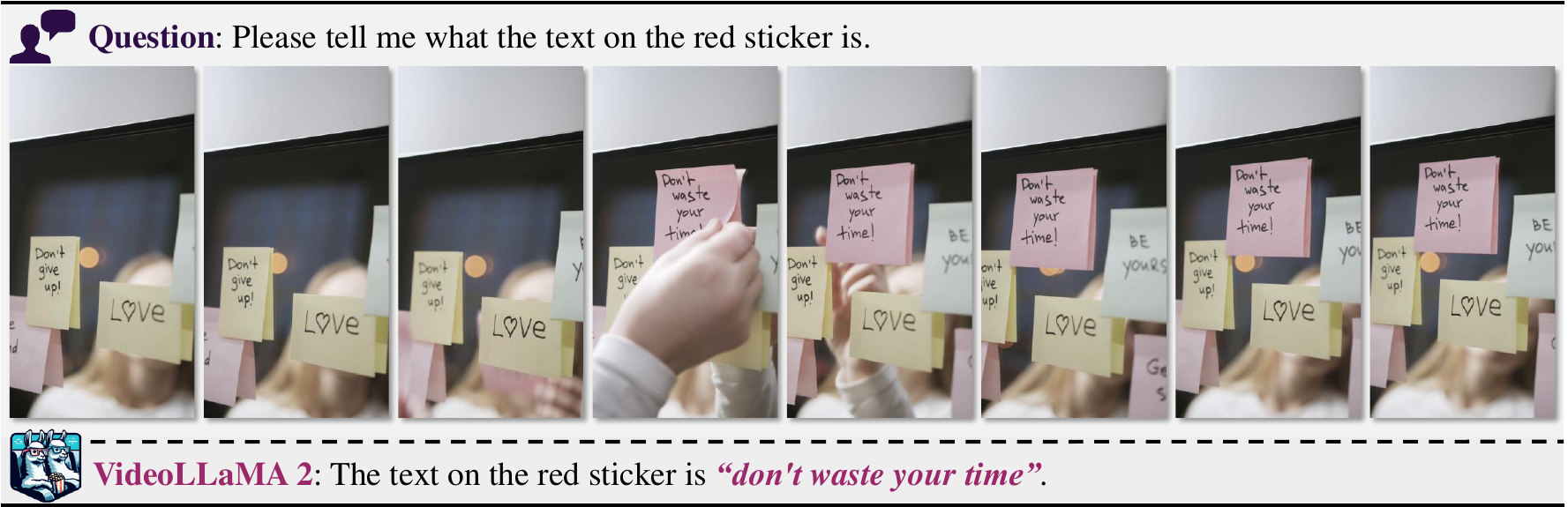}
    \caption{\textbf{Spatial-temporal Fine-grained Recognition Case.}}
    \label{case:fgr}
\end{subfigure}
\caption{\textbf{Qualitative Video Understanding Cases} from different perspectives of VideoLLaMA 2.}
\label{fig:case}
\end{figure}

As illustrated in Figure~\ref{fig:case}, we show some cases to demonstrate VideoLLaMA 2's multi-modal instruction-following capability in video-centric conversations:

\one~{\bf{Global Scene Understanding}}: As shown in Figure.~\ref{case:gsd}, It can be observed that VideoLLaMA 2 clearly describes the details of objects in the scene, such as the game console, the atmosphere of the scene, and the boy's dancing movements, which demonstrates the strong scene understanding ability of VideoLLaMA2.

\two~{\bf{Spatial-temporal Orientation Awareness}}: In Figure.~\ref{case:stp}, VideoLLaMA 2, by observing the entire video, correctly judged the car's turning direction, proving that VideoLLaMA2 is able to perceive spatial-temporal orientation information of a video.

\three~{\bf{Commonsense Reasoning}}: In Figure.~\ref{case:csr}, VideoLLaMA 2 first understands the events occurring in the video~(i.e., reading and drinking), then observes the environmental details in the video~(e.g., natural light), and finally infers that it is likely morning based on common sense. This demonstrates its strong capability for common sense reasoning.

\four~{\bf{Spatial-Temporal Fine-grained Recognition}}: 
Figure.~\ref{case:fgr} exhibits VideoLLaMA 2's fine-grained understanding ability of objects that appear at specific moments and specific locations in the video. Without being disturbed by the early information of the video, VideoLLaMA 2 accurately pinpoints the moment and location where the red sticky note appeared and recognizes the text on it.


\section{Related Works}

Existing Video-LLMs typically consist of a pre-trained visual encoder (e.g., CLIP~\citep{radford2021learning} or DINO~\citep{caron2021emerging}) to encode video frames into low-dimensional visual features, a vision-language adapter to aggregate these features and transform them into representations understandable by large language models (LLMs), and an instruction-tuned language decoder (e.g., LLaMA-2-Chat~\citep{touvron2023llama2} or Mistral-Instruct~\citep{jiang2023mistral}) to generate text responses based on instructions and user-uploaded videos. In this project, we primarily explore the better designs of vision-language connectors for Video-LLMs and leave the visual encoder and language decoder untouched.

The vision-language adapters of Image-LLMs, such as Cross-Attention~\citep{alayrac2022flamingoav,bai2023qwen}, Q-Former~\citep{li2023blip2bl,zhu2023minigpt,ye2024mplug}, 
Linear Projection~\citep{liu2023visualit,chen2023shikra,wang2023cogvlm} and
Dynamic Visual Tokenizer~\citep{jin2023unified}, have been widely adopted in popular Video-LLMs~\citep{zhang2023video,maaz2023videochatgpt,lin2023video,li2023mvbench,jin2024videolavit,ataallah2024minigpt4,liu2024st}. Despite their reasonable results, these designs are still insufficient for Video-LLMs because they completely ignore temporal aggregation and leave everything of temporal modeling to language decoder, which is neither computationally efficient (i.e., LLMs need to process much more tokens) nor effective on video understanding tasks, as observed in our preliminary experiments (See Table~\ref{tab:tc_abl}). 

Considering efficiency and effectiveness, we introduce the Spatial-Temporal Convolution Connector (STC Connector) into VideoLLaMA 2 to better capture the spatial-temporal features while maintaining a reasonable number of visual tokens. Specifically, our ST-Connector employs 3D convolution for spatial-temporal aggregation. To preserve local visual patterns during spatial compression, we follow~\citet{cha2023honeybee} to incorporate a RegStage~\citep{radosavovic2020designing} block before and after the 3D convolution, which has been shown to enhance spatial understanding.

Furthermore, expanding on the core capabilities of Video-LLMs, recent advancements have focused on integrating audio streams, a rich source of contextual cues that are vital for a complete video scene understanding. Models such as PandaGPT \citep{pandagpt} and Video-LLaMA \citep{zhang2023video} utilize pre-trained systems like ImageBind \citep{girdhar2023imagebind} for a universal modality alignment, while MacawLLM \citep{lyu2023macaw} optimizes separate branches for visual and audio data using diverse instructional sets. X-InstructBLIP (X-BLIP)  \citep{panagopoulou2023x} integrates multiple modalities within a single framework using frozen Large Language Models (LLMs), with modality-specific Q-Formers serving as adapters to bridge various encoders. OneLLM \citep{han2024onellm} introduces a universal encoder and projection module designed to align multiple modalities with linguistic data, thereby improving the coherence of multimodal integration. CREMA \citep{yu2024crema} takes an efficient and modular approach, utilizing parameter-efficient adapters for each modality to enhance flexibility and ease of incorporating new modalities into existing frameworks. Other models like AV-LLM \citep{shu2023audio} and AVicuna \citep{tang2024avicuna} leverage integrated audio-visual training data to further refine their understanding of complex multimodal content. These developments represent significant strides toward creating more versatile and capable LLMs that can navigate and interpret intricate temporal dynamics with synchronous audio streams effectively. 

\clearpage

\section{Conclusion}
We present the VideoLLaMA 2 series, a set of generalist Video Large Language Models designed to advance multimodal research in the arena of video-language modeling. By incorporating a meticulously designed \textit{Spatial-Temporal Convolution (STC) connector} and a jointly trained \textit{Audio Branch}, VideoLLaMA 2 consistently improves multimodal comprehension across various video and audio-oriented tasks. It outperforms the open-source models of similar size across multiple benchmarks and, in several aspects, achieves performance levels comparable to proprietary models, demonstrating impressive capabilities in modeling temporal dynamics with synchronous audio streams. Furthermore, as a foundational model, VideoLLaMA 2 can be further developed to benefit various more specialized but challenging problems, like long video understanding~\citep{ren2023timechat,song2023moviechat,wang2024videotree}, video agent~\citep{lin2023mm,fan2024videoagent,wang2024videoagent,he2024malmm}, autonomous driving~\citep{xu2023drivegpt4,shao2024accidentblip2}, motion understanding~\citep{wu2024motionllm,chen2024motionllm}, and robotic manipulation~\citep{liu2024enhancing}. 


\clearpage
\bibliography{neurips_2023}
\bibliographystyle{neurips_2023}

\newpage

\appendix

\section{The Prompt for GPT-aided AVQA Task Evaluation}\label{sec:gpt_eval}

\begin{table}[h]
\centering      
\begin{tabular}{p{13.5cm}}
\toprule
\textbf{System:} 
\texttt{You are an intelligent chatbot designed for evaluating the correctness of generative outputs for question-answer pairs. Your task is to compare the predicted answer with the correct answer and determine if they match meaningfully. Here's how you can accomplish the task:------\#\#INSTRUCTIONS: } \\
\texttt{- Focus on the meaningful match between the predicted answer and the correct answer.} \\
\texttt{- Consider synonyms or paraphrases as valid matches. } \\
\texttt{- Evaluate the correctness of the prediction compared to the answer.} \\
\hline
\textbf{User:}
\texttt{Please evaluate the following video-based question-answer pair:} \\
\texttt{Question: \textbf{"\{QUESTION\}"}} \\
\texttt{Correct Answer: \textbf{"\{ANSWER\}"}} \\
\texttt{Predicted Answer: \textbf{"\{PREDICTION\}"}} \\
\texttt{Provide your evaluation only as a yes/no and score where the score is an integer value between 0 and 5, with 5 indicating the highest meaningful match. Please generate the response in the form of a Python dictionary string with keys 'pred' and 'score', where value of 'pred' is  a string of 'yes' or 'no' and value of 'score' is in INTEGER, not STRING. DO NOT PROVIDE ANY OTHER OUTPUT TEXT OR EXPLANATION. Only provide the Python dictionary string. For example, your response should look like this: \{'pred': 'yes', 'score': 4\}.}\\
\bottomrule
\end{tabular}
\vspace{3pt}
\caption{The prompt for GPT-aided AVQA task evaluation.}
\end{table}

\section{Comparision Results with LLaVA-NeXT-Video Series}
\label{sec:comp_llava_next_video_dpo}

\begin{table}[h]
\centering                         
\renewcommand{\arraystretch}{1.4}  
\setlength{\tabcolsep}{0.5mm}      
\scriptsize                        
\begin{tabular}{y{100}x{30}|x{40}x{50}x{35}x{40}|x{40}x{40}}
\thickhline
\multirow{3}{*}{\bf{Model}} & \multirow{3}{*}{\bf{\# Frames}} & \multicolumn{4}{c|}{\bf{MC-VQA}} & \multicolumn{2}{c}{\bf{VC}} \\
\cline{3-8}
& &
\textbf{EgoSchema} &
\textbf{Perception-Test} &
\textbf{MVBench} & \textbf{VideoMME} & 
\multicolumn{2}{c}{\textbf{MSVC~(Score)}} \\
& & (Acc.) & (Acc.) & (Acc.) & (Acc.) & correctness & detailedness \\
\hline
LLaVA-NeXT-Video (7B) & 32 & 43.9\reproduce & 48.8\reproduce & 46.5\reproduce & 33.7\reproduce & 2.40\reproduce & 2.52\reproduce \\
LLaVA-NeXT-Video-DPO (7B) & 32 & 44.6\reproduce & 49.3\reproduce & 46.0\reproduce & 35.6\reproduce & \textbf{2.87}\reproduce & \textbf{2.86}\reproduce \\
\hdashline
VideoLLaMA 2 (7B)     & 8  & 50.5 & 49.6 & \multicolumn{1}{c}{53.4} & 44.0 & 2.57 & 2.61 \\
VideoLLaMA 2 (7B)     & 16  & 51.7 & 51.4 & \multicolumn{1}{c}{54.6} & 46.6 & 2.53 & 2.59 \\
VideoLLaMA 2 (8x7B)   & 8  & 53.3 & \textbf{52.2} & \multicolumn{1}{c}{53.9} & \textbf{48.4} & 2.53 & 2.56 \\
\thickhline
\end{tabular}
\vspace{3pt}
\caption{Comparison results on MV-VQA and VC tasks. \textcolor{darkpurple}{$\spadesuit$}: reproduced results.}
\label{tab:llava-next-video-dpo_mcvqa_vc}
\end{table}

\begin{table}[h]
\centering                         
\renewcommand{\arraystretch}{1.4}  
\setlength{\tabcolsep}{0.5mm}      
\scriptsize                        
\begin{tabular}{y{100}x{30}|x{40}x{40}|x{32}x{32}x{32}x{32}x{35}}
\thickhline
\multirow{2}{*}{\textbf{Model}} & \multicolumn{1}{c|}{\multirow{2}{*}{\textbf{\# Frames}}} & \bf{MSVD} & \bf{ActivityNet} & \multicolumn{5}{c}{\textbf{Video-ChatGPT~(Score)}} \\
\cline{3-9} 
& & (Acc./Score) & (Acc./Score) & Correctness & Detail & Context & Temporal & Consistency \\ 
\hline
LLaVA-NeXT-Video (7B) & 32 & 67.8/3.5\reproduce & 53.5/3.2\official & 3.39\official & 3.29\official & 3.92\official & 2.60\official & 3.12\official \\
LLaVA-NeXT-Video-DPO (7B) & 32 & 71.0/3.7\reproduce & \textbf{60.2/3.5}\official & \textbf{3.64}\official & \textbf{3.45}\official & \textbf{4.17}\official & \textbf{2.95}\official & \textbf{4.08}\official  \\
\hdashline
VideoLLaMA 2 (7B)    & 8  & \textbf{71.7/3.9} & 49.9/3.3 & 3.09 & 3.09 & 3.68 & 2.63 & 3.25 \\
VideoLLaMA 2 (7B)    & 16 & 70.9/3.8 & 50.2/3.3 & 3.16 & 3.08 & 3.69 & 2.56 & 3.14 \\
VideoLLaMA 2 (8x7B)  & 8  & 70.5/3.8 & 50.3/3.4 & 3.08 & 3.11 & 3.64 & 2.67 & 3.26 \\
\thickhline
\end{tabular}
\vspace{3pt}
\caption{Comparison results on MSVD, ActivityNet, and Video-ChatGPT. \textcolor{plum}{$\varheartsuit$}: officially reported results. {\textcolor{darkpurple}{$\spadesuit$}}: reproduced results.
}
\label{tab:llava-next-video-dpo_oevqa}
\end{table}

\newpage
\section{Results on More Video Benchmarks}
\label{app:additional results}
\begin{table}[h]
\centering                         
\renewcommand{\arraystretch}{1.4}  
\setlength{\tabcolsep}{0.5mm}      
\scriptsize                        
\begin{tabular}{y{80}x{30}|x{40}x{40}x{60}x{50}}
\thickhline
\multirow{2}{*}{\textbf{Model}} & \multicolumn{1}{c|}{\multirow{2}{*}{\textbf{\# Frames}}} & \bf{MLVU} & \bf{VideoVista} & \bf{MMBench-Video} & \bf{DREAM-1k} \\
\cline{3-6}
& & (M-AVG) & (Acc.) & (Score) & (F1) \\ 
\hline
VideoLLaMA 2 (7B)    & 16 & 32.7 & 60.47 & 1.08 & 26.2 \\
VideoLLaMA 2 (72B)   & 16 & 45.6 & -     & -    & 27.1 \\
\thickhline
\end{tabular}
\vspace{4pt}
\caption{Results of VideoLLaMA 2 models on MLVU~\citep{MLVU}, VideoVista~\citep{li2024videovista}, MMBench-Video~\citep{fang2024mmbenchvideo}, and DREAM-1k~\citep{wang2024tarsierrecipestrainingevaluating} benchmarks.}
\end{table}
\end{document}